\documentclass[a4paper]{article}

\usepackage{INTERSPEECH2022}
 
 \usepackage{cite}
 
\usepackage{hyperref}

\usepackage{array,booktabs,longtable,tabularx,tabulary}

\usepackage{microtype}
\usepackage[dvipsnames]{xcolor}
\usepackage[T1]{fontenc}
\usepackage[utf8]{inputenc}

\usepackage{booktabs}

\usepackage{booktabs}
\usepackage{multirow}

\title{Integrating Form and Meaning: \\ A Multi-Task Learning Model for Acoustic Word Embeddings}
\name{Badr M. Abdullah, Bernd M\"obius, Dietrich Klakow}
\address{
Language Science and Technology (LST), Saarland University, Germany \\
Saarland Informatics Campus, Germany
  }
\email{\{babdullah|moebius|dietrich\}@lsv.uni-saarland.de}

\begin{document}

\maketitle
\begin{abstract}

\end{abstract}
Models of acoustic word embeddings (AWEs) learn to map variable-length spoken word segments onto fixed-dimensionality vector representations such that different acoustic exemplars of the same word are projected nearby in the embedding space. 
In addition to their speech technology applications, AWE models have been shown to predict human performance on a variety of auditory lexical processing tasks. 
Current AWE models are based on neural networks and trained in a bottom-up approach that integrates acoustic cues to build up a word representation given an acoustic or symbolic supervision signal. 
Therefore, these models do not leverage  or capture high-level lexical knowledge during the learning process. 
In this paper, we propose a multi-task learning model that incorporates top-down lexical knowledge into the training procedure of AWEs. 
Our model learns a mapping between the acoustic input and a lexical representation that encodes high-level information such as word semantics in addition to bottom-up form-based supervision. 
We experiment with three languages and demonstrate that incorporating lexical knowledge improves the embedding space discriminability  and encourages the model to better separate lexical categories.

 

\section{Introduction}

The development of robust automatic speech recognition (ASR) systems requires large collections of high-quality transcribed speech, which are only available for a small subset of the world languages. 
To facilitate access to spoken content for language varieties that are not yet supported by conventional ASR systems, researchers have developed voice-based search applications such as query-by-example (QbE) search~\cite{zhang2009unsupervised,jansen2012indexing,metze2013spoken}. These systems rely on vector-space acoustic models that map variable-length spoken word segments onto fixed-size vector representations such that exemplars of the same word are (ideally) projected onto the same vector~\cite{levin2013fixed,bengio2014word,kamper2016deep,settle2016discriminative,settle2017query}. 
In the speech technology literature, these fixed-dimensionality vector representations are known as acoustic word embeddings (AWEs). 
Currently, the top performing and the most efficient models of AWEs are based on deep neural networks (DNNs) \cite{kamper+etal_icassp16, he+etal_iclr17, kamper2015unsupervised, kamper2019truly}.
Due to the ubiquity of computers that support DNNs coupled with highly-optimized vector-space search algorithms \cite{JDH17}, AWEs enable efficient indexing and retrieval of spoken content at an unprecedented scale.

In addition to their applications in speech technology, DNN-based models of AWEs have been adopted as models of human speech processing and analyzed from a cognitively motivated angle in recent studies. 
For example, it has been shown that AWEs  exhibit a  human-like word onset bias where distinct words are more likely to be perceived as similar if they begin with the same sound \cite{matusevych+etal_baics20}.
Furthermore, AWE models have been shown to predict non-native perceptual difficulties in phonetic categorization \cite{MatusevychSKFG20}, cross-linguistic effects in auditory lexical processing \cite{matusevych2021phonetic}, and the facilitation effect of cross-language similarity on spoken-word processing  \cite{abdullah-etal-2021-familiar, mayn-etal-2021-familiar}.
In non-native word production, models of AWEs have been reported to capture lexical production patterns of second language (L2) learners \cite{ando21_interspeech}. 
These empirical findings from cognitively motivated, computational word perception and production studies encourage further integration between speech technology and cognitive science. 


\begin{figure}
    \centering
     \includegraphics[width=0.40\textwidth]{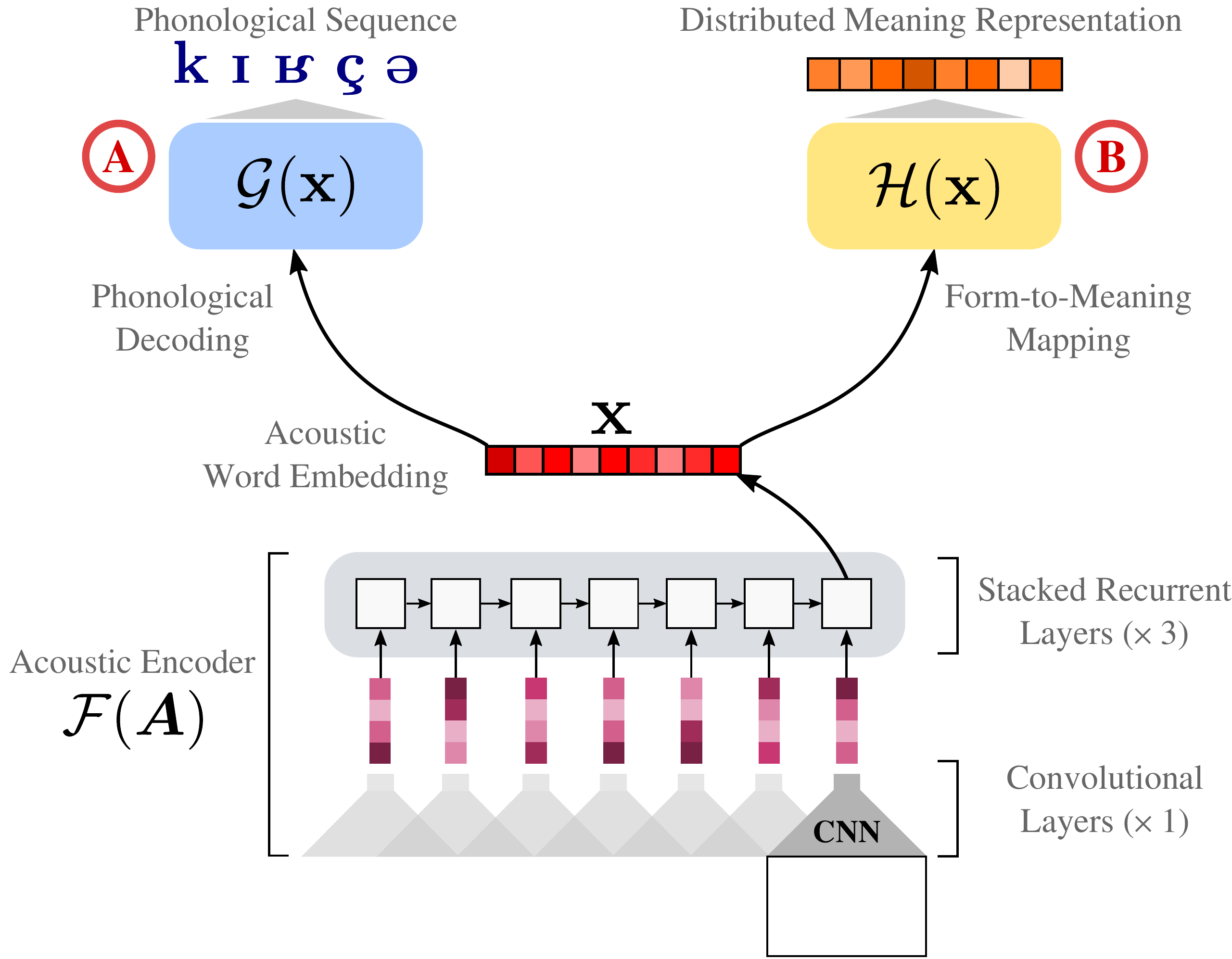}
    \caption{A schematic view of our proposed model. }
    \label{fig:model}
\end{figure}

Nevertheless, the majority of existing AWEs rely on supervision signals that only capture low-level, form-based information about the word.
That is, AWEs are learned in a bottom-up approach whereby acoustic-phonetic cues are integrated in the model to build up a word form representation that encodes its phonetic features and phonological structure.
However, a host of psycholinguistic studies with human listeners have shown that top-down, high-level lexical properties---such as word semantics---not only interact with the word recognition process but also facilitate discrimination between word competitors \cite{zhuang2011interaction, mirman2009dynamics, strain1995semantic, cortese1997effects, hino1996effects}.
We take inspiration from these experimental findings and introduce an AWE model based on the multi-task learning framework that integrates form-based and meaning-based supervision signals into a single model (Fig. 1). Contrary to prior work that aims to learn the semantic content directly using a very large speech corpus \cite{ChungG18}, our model incorporates word semantics as an additional supervision signal, thus requiring only a few hours of speech and being more applicable in low-resource settings. 
We experiment with read speech corpora for three languages and empirically demonstrate that integrating high-level lexical knowledge into training AWEs improves the ability of the model to discriminate between lexical categories.

\newpage

\section{AWEs via Multi-Task Learning}
Given an acoustic signal that corresponds to a spoken word represented as a temporal sequence of $T$ spectral vectors, i.e., ${\boldsymbol{A}} = (\boldsymbol{a}_1, \boldsymbol{a}_2, ..., \boldsymbol{a}_T)$, the goal of an AWE model is to transform $\boldsymbol{A}$ into a $D$-dimensional vector representation $\mathbf{x}$. 
This task corresponds to learning an encoder function $\mathcal{F}_{\boldsymbol{\theta}}: \mathcal{A} \xrightarrow[]{} \mathbb{R}^D$, where $\mathcal{A}$ is the (continuous) space of acoustic sequences, $D$ is the embedding dimensionality, and  $\boldsymbol{\theta}$ are the parameters of the function. 
Sequences in $\mathcal{A}$ can vary in length, thus the function $\mathcal{F}_{\boldsymbol{\theta}}$ should be modeled with a suitable neural architecture such as recurrent networks (RNNs). 
Therefore, transforming a variable-length acoustic input into a $D$-dimensional AWE can be described as 
\begin{equation}
\mathbf{x} = \mathcal{F}(\boldsymbol{A}; \boldsymbol{\theta}_\mathcal{F}) \in \mathbb{R}^D
\end{equation}
Different approaches in the literature have been proposed for modeling the function $\mathcal{F}({.; \boldsymbol{\theta}_\mathcal{F})}$, which can be characterized as either architectural innovations or introducing new loss functions.
In the approach we propose in this paper, which is inspired by a classical connectionist model of speech processing \cite{gaskell1997integrating}, our goal is to integrate two sources of supervision signals---namely phonological form and lexical semantics---into the training procedure. 
To this end, we assume a dataset  $\mathcal{D} = \{(\boldsymbol{A}^1, w^1), \dots, (\boldsymbol{A}^N, w^N)\}$ of $N$ spoken words where $w^{i}$ is the written form of the $i$th word. 
Such a dataset can be automatically obtained using a forced alignment tool on a transcribed speech dataset. 
Furthermore, we assume the availability of two look-up dictionaries: (1) a dictionary that maps each written word onto its phonetic transcription as $\Phi(w) = \boldsymbol{\varphi}_{1:\tau} = \{\varphi_1, \dots, \varphi_{\tau}\}$, which can be automatically created using a grapheme-to-phoneme (G2P) tool,  and (2) a lookup dictionary that maps each word into a distributed word representation as $\Lambda(w) = \mathbf{w} \in  \mathbb{R}^K$.
The distributed word representation ideally encodes high-level lexical knowledge about the word---such as its semantic and syntactic properties---and can be obtained independently using a large text corpus or from a public repository of semantic word embeddings such as  \textbf{\textit{Glove}} \cite{pennington2014glove} or \textbf{\textit{fasttext}} \cite{mikolov-etal-2018-advances}.

\subsection{Form-based Phonological Supervision}
Our first  learning objective is based on the sequence-to-sequence learning framework in which the network is trained as a word-level acoustic model (Fig.\ref{fig:model}, branch [A]).
Given the output of acoustic encoder $\mathbf{x}$, a phonological decoder $\mathcal{G}(.; \boldsymbol{\theta}_\mathcal{G})$ aims to decode the corresponding phonological sequence $\boldsymbol{\varphi}_{1:\tau}$ of the word form $\mathbf{x}$.
 The objective is to minimize a categorical cross-entropy loss at each timestep in the decoder, which is equivalent to minimizing the term
\begin{equation}
    \begin{aligned}
        \mathcal{L}^\phi(\boldsymbol{\theta}_\mathcal{F}, \boldsymbol{\theta}_\mathcal{G}) & = - \sum_{({\boldsymbol{A}}^i, w^i) \in \mathcal{D}} \: { \text{log } \mathbf{P}\big({\Phi}(w^i)\: | \: \mathbf{x}^i ; \: \boldsymbol{\theta}_\mathcal{G}\big)} \\
        & = - \sum_{({\boldsymbol{A}}^i, w^i) \in \mathcal{D}} \: \sum_{t = 1}^{\tau}{ \text{log } \mathbf{P}\big(\varphi_{t} \: | \: t, \: \mathbf{x}^i ; \: \boldsymbol{\theta}_\mathcal{G}\big)}
    \end{aligned}
\end{equation} 
where $\mathbf{P}\big(\varphi_{t} \: | \: t, \: \mathbf{x}^i ; \: \boldsymbol{\theta}_\mathcal{G}\big)$ is the probability of the phoneme $\varphi_{t}$ at the $t$th timestep, conditioned on the previous phoneme sequence $\boldsymbol{\varphi}_{1:t-1}$ and the AWE $\mathbf{x}$, and $\boldsymbol{\theta}_\mathcal{G}$ are the parameters of the decoder. 
The intuition of this learning objective is the following: although their acoustic realizations vary due to speaker and context variability, different exemplars of the same word category would have identical phonetic transcriptions.
Therefore, we expect the model to project exemplars of the same lexical category nearby in the space and embedding distance should ideally correlate with phonological (dis)similarity.


\subsection{Meaning-based Lexical Supervision}

Our second learning objective aims to map the acoustic input $\boldsymbol{A}$ onto a high-level lexical representation  (Fig.\ref{fig:model}, branch [B]). 
The goal here is to incorporate a supervision signal from a  level that is higher in the linguistic hierarchy compared to form-based phonological  supervision. 
Inspired by Maas et al. \cite{maas2012word}, we model this task as a vector regression problem. 
The output of the acoustic encoder $\mathbf{x}$ is transformed via a feed-forward network into a semantic vector as $\boldsymbol{v} = \mathcal{H}(\mathbf{x}; \boldsymbol{\theta}_{\mathcal{H}})\in \mathbb{R}^K$. Thus, the objective is to minimize the term 
\begin{equation}
    \mathcal{L}^\lambda(\boldsymbol{\theta}_\mathcal{F}, \boldsymbol{\theta}_\mathcal{H}) = \sum_{({\boldsymbol{A}}^i, w^i) \in \mathcal{D}} \:  || \: \boldsymbol{v}^i - \Lambda(w^i) \:||_2
\end{equation}  
where $\Lambda(w^i) \in \mathbb{R}^K$ is the ground-truth distributed representation, or semantic word embedding, of the $i$th sample.
We assume that continuous, distributed word representations are available to the model during training. 
Given the ubiquity of word embeddings in the NLP research and the availability of text corpora for many languages, we believe that our assumption is reasonable.   

\subsection{Integrating Form and Meaning Supervision}
To integrate the two sources of supervision when training the model, we jointly minimize the term 

\begin{equation}
     \mathcal{L}(\boldsymbol{\theta}_\mathcal{F}, \boldsymbol{\theta}_\mathcal{G}, \boldsymbol{\theta}_\mathcal{H}) = \alpha \cdot \mathcal{L}^\phi + \beta \cdot \mathcal{L}^\lambda
\end{equation}
Here, $\alpha$ and $\beta$ are trade-off hyperparameters (i.e., scalars) that control the contribution of each term to the overall loss.

\section{Baseline: Contrastive Acoustic Model}
We compare the performance of our proposed model to a strong baseline that explicitly minimizes the distance between exemplars of the same lexical category.  
The baseline model employs a contrastive triplet loss that has been extensively explored in the AWEs literature with different underlying architectures and has shown strong discriminative performance \cite{settle+livescu_slt16, kamper+etal_icassp16, Abdullah2021DoAW, Jacobs2021MultilingualTO}. Given a matching pair of AWEs $(\mathbf{x}^a, \mathbf{x}^+)$---i.e., embeddings of two exemplars of the same word type---the objective is then to minimize a triplet margin loss
\begin{equation}
    \mathcal{L}(\boldsymbol{\theta}_\mathcal{F}) = \sum_{({\boldsymbol{A}}^i, w^i) \in \mathcal{D}} \: \text{max} \big[0, m + d(\mathbf{x}^i, \mathbf{x}^+) - d(\mathbf{x}^i, \mathbf{x}^-) \big]
\end{equation}
where $\mathbf{x}^-$ is an AWE that corresponds to a word other than $w^i$, and $d: \mathbb{R}^D \times \mathbb{R}^D \rightarrow [0, 1]$ is the cosine distance. This objective aims to map acoustic exemplars of the same word closer in the embedding space while pushing away segments of different word types by a distance defined by the margin hyperparameter $m$. To obtain negative samples, we create mismatching pairs from the mini-batch such that  $d(\mathbf{x}^i, \mathbf{x}^-)$ is minimized \cite{jansen2018unsupervised}.


\section{Experiments}

\begin{figure*}[t!]
    \centering
     \includegraphics[width=0.95\textwidth]{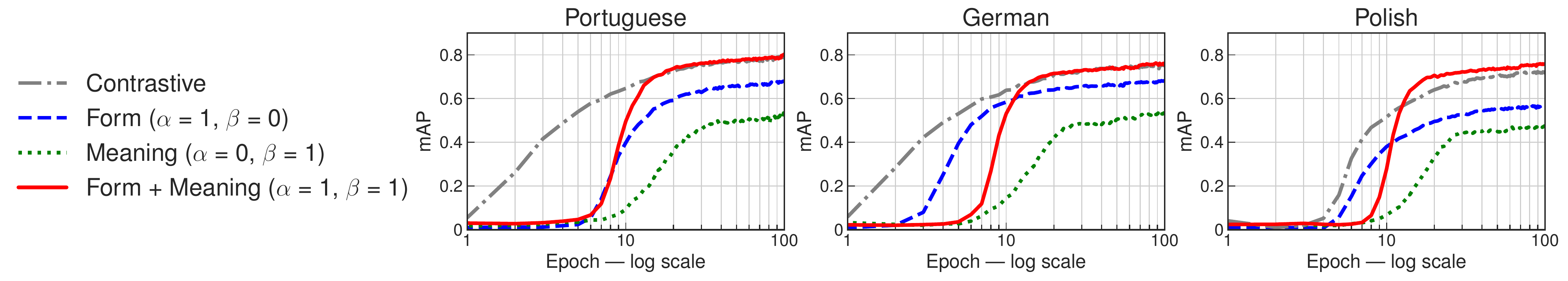}
    \caption{Learning curves of the models for 100 training epochs, quantified by the word discrimination task and the mAP metric.}
    \label{fig:learning_curve}
\end{figure*}

\subsection{Experimental Data}

\begin{table}[b]
\centering
\caption{Word-level statistics of our experimental data.}
\label{tab:data-stats}
\begin{tabular}{@{}cccccc@{}}
\toprule
\multirow{2}{*}{} & \multicolumn{3}{c}{\# segments per split} & \multirow{2}{*}{\begin{tabular}[c]{@{}c@{}}duration\\ ($\mu \pm \sigma$)\end{tabular}} & \multirow{2}{*}{\begin{tabular}[c]{@{}c@{}}TTR\end{tabular}} \\ \cmidrule(lr){2-4}
 & train & valid & test &  &  \\ \midrule
 Portuguese & 28810 &  9029 & 9580  & 0.51 $\pm$ 0.19 & 0.147 \\
German & 28914  & 9683 & 9372 & 0.44 $\pm$ 0.18 & 0.193 \\
Polish & 27979 & 9656 & 9089 & 0.50 $\pm$ 0.18 & 0.267 \\ \bottomrule
\end{tabular}
\end{table}


The data in our study is drawn from the GlobalPhone multilingual speech database \cite{schultz2013globalphone} for Portuguese, German, and Polish (see Table \ref{tab:data-stats}). 
We sample 42 speakers from each language for training and obtain  spoken word segments  using the Montreal Forced Aligner \cite{mcauliffe2017montreal}. 
It is worth pointing out that the speakers in the validation and test splits are held-out and not used while training.
The phonetic transcription for each word is produced using the \textit{eSpeak}  G2P tool.  
Then, each acoustic segment is parametrized  as a sequence of 39-dimensional Mel-frequency spectral coefficients of 25ms frames with 10ms overlap. 


\subsection{Architecture and Hyperparameters}
\textbf{Acoustic Encoder $\mathcal{F}(.; \boldsymbol{\theta}_{\mathcal{F}})$.} \hspace{0.1cm} Our acoustic encoder consists of a hybrid, convolutional-recurrent neural network architecture.
The frond-end consists of a 1D convolutional layer of 64 filters with a kernel size of 5 spectral vectors and stride of 2. Then, the output of the convolutional layer is fed sequentially into a recurrent block that consists of a 3-layer unidirectional Gated Recurrent Unit (GRU) with a hidden state of 512 units, which yields  a 512-dimensional AWE as the last hidden state of the GRU. We apply layer-wise dropout with a probability of 0.2. Bidirectional GRUs did not yield further improvements.
\vspace{0.1cm}

\noindent
\textbf{Phonological Decoder $\mathcal{G}(.; \boldsymbol{\theta}_{\mathcal{G}})$.} \hspace{0.1cm} We employ a 1-layer GRU of 512 units hidden state that takes the 512-dimensional AWE as the initial hidden state and decodes the corresponding phonological sequence without teacher forcing. 
\vspace{0.1cm}

\noindent
\textbf{Form-to-Meaning Regressor $\mathcal{H}(.; \boldsymbol{\theta}_{\mathcal{H}})$.}  \hspace{0.1cm} We employ a linear layer (512 $\rightarrow$ 300)  followed by a \textbf{\texttt{tanh}} non-linearity to project the AWE $\mathbf{x}$  onto the corresponding distributed word representation. We use pre-trained 300-dimensional fasttext embeddings as  distributed word representations. Deeper feed-forward networks did not yield further improvements. 
\vspace{0.1cm}

\noindent
\textbf{Contrastive Loss.} \hspace{0.1cm} For the baseline model with the contrastive loss, we experiment with different values of the margin hyperparameter $m = \{0.2, 0.3, 0.4, 0.5\}$, out of which $0.4$ yields the best performance on the validation set.
\vspace{0.1cm}

\noindent
\textbf{Training Details.} \hspace{0.1cm} We train all models in this study for 100 epochs with batches of 256 samples using the Adam optimizer \cite{DBLP:journals/corr/KingmaB14} with an initial learning rate (LR) of 0.001. The LR is reduced  by a factor of 0.5 if the performance on the validation set does not improve for 10 epochs. The epoch with the best validation performance is used for evaluation on the test set.
\vspace{0.1cm}

\noindent
\textbf{Implementation.} \hspace{0.1cm} We develop our code using PyTorch \cite{paszke2019pytorch} and we make it publicly available (\small{\url{https://github.com/uds-lsv/semantically_enriched_AWEs}}). 

\begin{figure}
    \centering
     \includegraphics[width=0.45\textwidth]{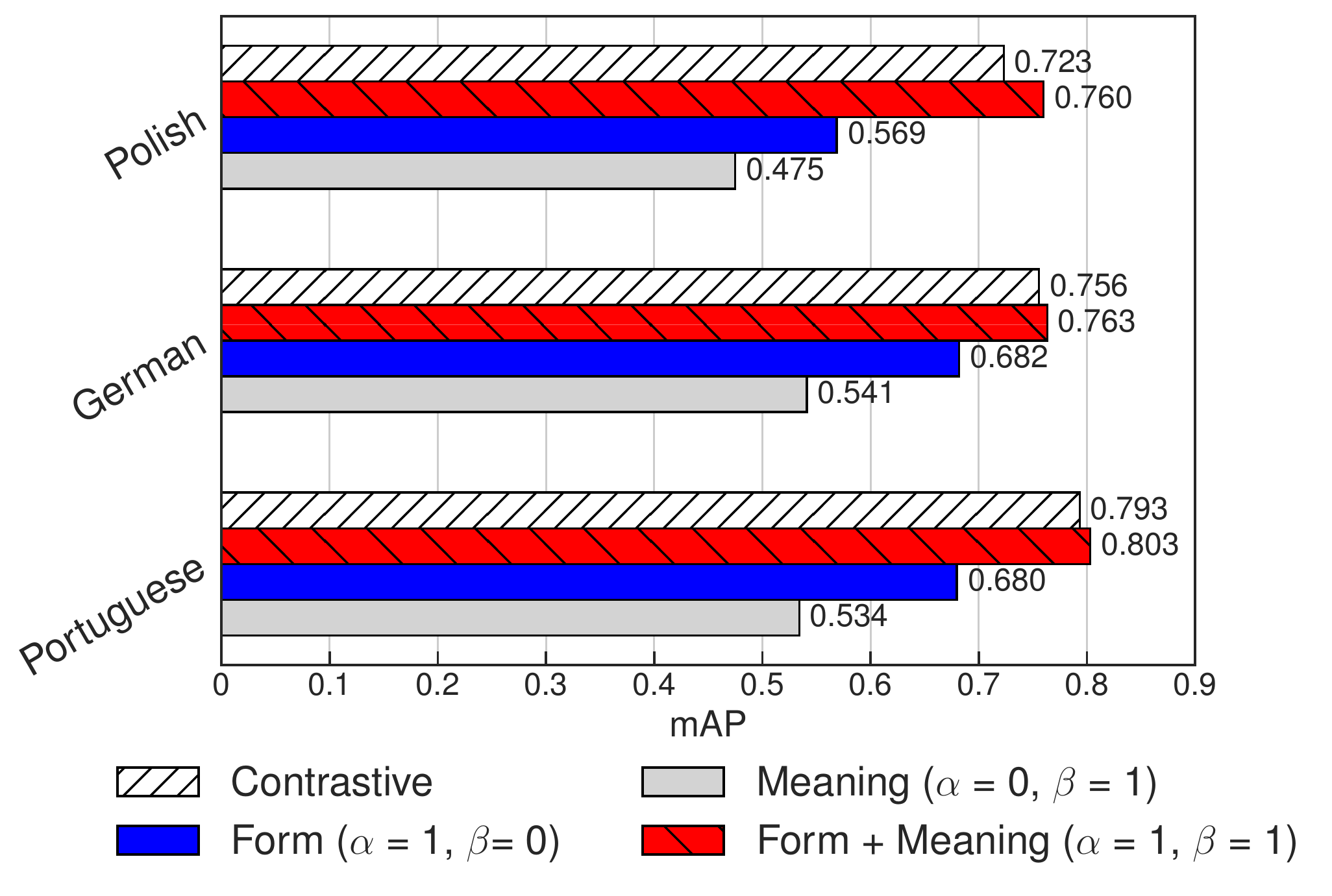}
    \caption{Word discrimination performance (mAP) on test set. }
    \label{fig:best_mAP}
\end{figure}

\subsection{Experimental Results}
We conduct an intrinsic evaluation for the AWEs to assess the performance of our models using the same-different acoustic word discrimination task with the mean average precision (mAP) metric \cite{algayres20_interspeech, Abdullah2021DoAW, settle+etal_icassp19}.  Prior work has shown that performance on this task positively correlates with improvement on downstream QbE speech search \cite{Jacobs2021MultilingualTO}.  This task evaluates the ability of the model to determine whether two given speech segments correspond to the same word type---that is, whether or not two acoustic segments are exemplars of the same category.  

Fig.~\ref{fig:learning_curve} shows the learning curves for the models during 100 epochs of training quantified by the performance on the validation set.
Contrary to the other models, we observe that the contrastive baseline model reaches a reasonable performance before the $10$th epoch, which we attribute to the fact that the evaluation task (word discrimination) and the learning objective (contrastive triple loss) are analogous.
Fig.~\ref{fig:best_mAP} shows the final performance on the test set. 
We observe that both the form-only model ($\alpha = 1, \beta = 0$) and the meaning-only model ($\alpha = 0, \beta = 1$) perform poorly compared to the contrastive baseline. 
However, integrating the two sources of supervision in the form + meaning setting ($\alpha = 1, \beta = 1$) enables the model to outperform the contrastive baseline for the three languages in our study.  
The gain in performance is more prominent in the Polish language (relative mAP gain by $5.06\%$), which is the most morphologically complex language in our study due to its rich inflection system.
The Polish morphological complexity is also reflected in its relatively high type-to-token ratio (TTR) in Table~\ref{tab:data-stats}. 
These findings show that integrating high-level linguistic knowledge in training acoustic models improves the discriminability of the embeddings space, and the effect seems to be more prominent on a language with a rich morphological system.

\begin{figure}
    \centering
    \includegraphics[width=0.45\textwidth]{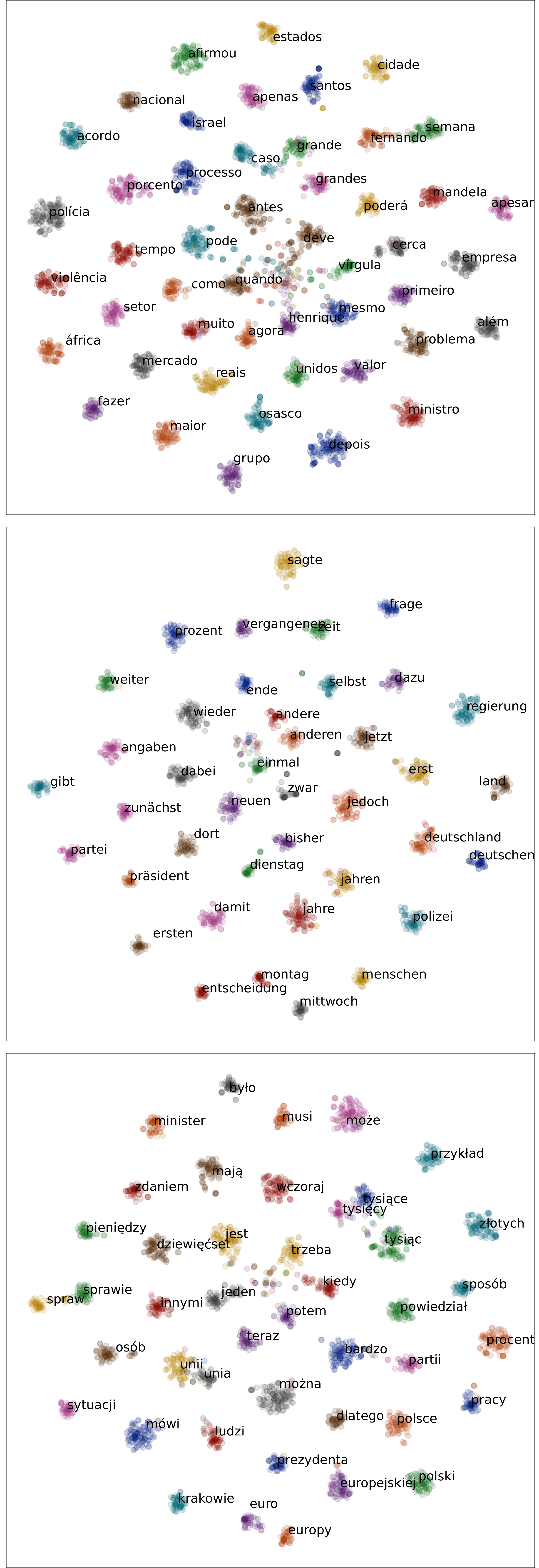}
    \caption{Two-dimensional visualization of the word embedding space using the t-SNE algorithm: (Top) Portuguese, (Middle) German, and (Bottom) Polish. Figure is best viewed in color.}
    \label{fig:tsne_viz}
\end{figure}

\newpage

\subsection{Embedding Visualization}

To analyze our multi-task learning model and gain further insights into its emergent embedding space, we use the t-SNE \cite{van2008visualizing} dimensionality reduction algorithm on the AWEs from the setting where $\alpha = 1$ and $\beta = 1$ (form + meaning supervision).
We visualize the embeddings of acoustic word samples from  held-out speakers (i.e., a set of speakers the models were not trained on). 
Note that the t-SNE objective aims to preserve the local structure within the higher-dimensional space when reducing the dimensionality.
Therefore, the local distance between the two-dimensional projections of the embeddings mainly reflects the cluster structure within the embedding space, which enables us to visually inspect the emergent clusters and investigate whether or not they correspond to distinct lexical categories.

The  t-SNE visualizations for the three languages in our study are illustrated in Fig.~\ref{fig:tsne_viz}.
We observe a clear tendency for exemplars of the same lexical category to closely cluster in the embedding space, despite the lack of an explicit clustering objective in the learning procedure. 
One notable exception we observe in Fig.~\ref{fig:tsne_viz} (bottom) for the Polish language is the two nearby, nearly overlapping clusters that correspond to the word forms [tysięcy] and [tysiące]. 
Note that these two word forms are two morphological variants of the same lemma (i.e., [tysiąc], the Polish word for \textit{thousand}). Given their semantic and phonological similarity, a small distance between the centroids of their clusters is expected. 

\section{Discussion and Conclusion}
AWEs are vector representations of spoken words that encode their acoustic-phonetic features and phonological structures.
In addition to their utility in speech technology applications, models of AWEs have shown to produce human-like behavior in various auditory lexical processing tasks. 
Existing methods for learning AWEs from speech corpora employ training strategies with acoustic, phonological feature-based, or symbolic form-based supervision. 
These learning strategies correspond to the bottom-up integration of acoustic-phonetic cues to build up a word form representation. In our paper, we have introduced a methodology based on the multi-task learning framework that leverages top-down, high-level lexical knowledge to learn semantically-enriched AWEs.
We have experimented with semantic word embeddings as distributed meaning representations that guide the learning process in addition to form-based phonological supervision. 
Our experiments have demonstrated that integrating the two sources of supervision (i.e., phonological form and lexical semantics) improves the discriminability of the embeddings space---for the three languages in our study---as evidenced by the competent performance of our model compared to a strong contrastive AWE model.
Furthermore, the t-SNE visualization analysis has supported our experimental findings in the word discrimination evaluation and provided further evidence that incorporating top-down lexical knowledge encourages the model to better separate the lexical categories in the embedding space without explicit supervision that directs the network to minimize the distance in the embedding space or a clustering objective.  

\section{Acknowledgements}
We thank the anonymous reviewers for their feedback and Miriam Schulz for her comments on the paper.
This research is funded by the Deutsche Forschungsgemeinschaft (DFG, German Research Foundation), Project-ID 232722074 -- SFB 1102.

\bibliographystyle{IEEEtran}
\bibliography{mybib}



\end{document}